\documentclass[11pt,a4paper]{article}
\usepackage[hyperref]{naaclhlt2019}
\usepackage{times}
\usepackage{subfig}
\usepackage{latexsym}
\usepackage{graphicx}
\usepackage{amsfonts}
\usepackage{amsmath}
\usepackage{amssymb}
\usepackage{todonotes}
\usepackage{url}

\aclfinalcopy 
\def\aclpaperid{1363} 

\title{Abusive Language Detection with Graph Convolutional Networks}

\author{Pushkar Mishra$^\bigstar$,\,\, Marco Del Tredici$^\clubsuit$,\,\, Helen Yannakoudakis$^\spadesuit$,\,\, Ekaterina Shutova$^\clubsuit$\\
  $^\bigstar$ Facebook AI, London, United Kingdom\\
  $^\clubsuit$ ILLC, University of Amsterdam, The Netherlands\\
  $^\spadesuit$ The ALTA Institute, Dept. of CS \& Technology, University of Cambridge, United Kingdom\\
  \normalsize{{\tt pushkarmishra@fb.com, m.deltredici@uva.nl, hy260@cl.cam.ac.uk, e.shutova@uva.nl}}
}

\date{}

\begin{document}
\maketitle
\begin{abstract}
Abuse on the Internet represents a significant societal problem of our time. Previous research on automated abusive language detection in Twitter has shown that community-based profiling of users is a promising technique for this task. However, existing approaches only capture shallow properties of online communities by modeling follower--following relationships. In contrast, working with graph convolutional networks (\textsc{gcn}s), we present the first approach that captures not only the structure of online communities but also the linguistic behavior of the users within them. We show that such a heterogeneous graph-structured modeling of communities significantly advances the current state of the art in abusive language detection.
\end{abstract}

\section{Introduction}
Matthew Zook \shortcite{zook} carried out an interesting study showing that the racist tweets posted in response to President Obama's re-election were not distributed uniformly across the United States but instead formed clusters. This phenomenon is known as \textit{homophily}: i.e., people, both in real life and online, tend to cluster with those who appear similar to themselves. To model homophily, recent research in abusive language detection on Twitter \cite{mishra} incorporates embeddings for \textit{authors} (i.e., users who have composed tweets) that encode the structure of their surrounding communities. The embeddings (called \textit{author profiles}) are generated by applying a node embedding framework to an undirected unlabeled community graph where nodes denote the authors and edges the follower--following relationships amongst them on Twitter. However, these profiles do not capture the linguistic behavior of the authors and their communities and do not convey whether their tweets tend to be abusive or not.

In contrast, we represent the community of authors as a heterogeneous graph consisting of two types of nodes, authors and their tweets, rather than a homogeneous community graph of authors only. The primary advantage of such heterogeneous representations is that they enable us to model both community structure as well as the linguistic behavior of authors in these communities. To generate richer author profiles, we then propose a semi-supervised learning approach based on graph convolutional networks (\textsc{gcn}s) applied to the heterogeneous graph representation. To the best of our knowledge, our work is the first to use \textsc{gcn}s to model online communities in social media. We demonstrate that our methods provide significant improvements over existing techniques.

\section{Related work}
Supervised learning for abusive language detection was first explored by Spertus \shortcite{smokey} who extracted rule-based features to train their classifier. Subsequently, manually-engineered lexical--syntactic features formed the crux of most approaches to the task \cite{yin,warner}. Djuric et al. \shortcite{djuric} showed that dense comment representations generated using \textit{paragraph2vec} outperform bag-of-words features. Several works have since utilized (deep) neural architectures to achieve impressive results on a variety of abuse-annotated datasets \cite{nobata,pavlopoulos}. Recently, the research focus has shifted towards extraction of features that capture behavioral and social traits of users. Pavlopoulos et al. \shortcite{W17-4209} showed that including randomly-initialized user embeddings improved the performance of their \textsc{rnn} methods. Qian et al. \shortcite{qian} employed \textsc{lstm}s to generate inter and intra-user representations based on tweets, but they did not leverage community information.

\section{Dataset}
\label{data}
Following previous work \cite{mishra}, we experiment with a subset of the Twitter dataset compiled by Waseem and Hovy \shortcite{waseem_hovy}. Waseem and Hovy released a list of $16,907$ tweet IDs along with their corresponding annotations,\footnote{\url{https://github.com/ZeerakW/hatespeech/}} labeling each tweet as \textit{racist}, \textit{sexist} or \textit{neither} \textit{(clean)}. Recently, Mishra et al. \shortcite{mishra} could only retrieve $16,202$ of these tweets since some of them are no longer available. This is the dataset we use in our experiments. $1,939$ ($12\%$) of $16,202$ tweets are \textit{racist}, $3,148$ ($19.4\%$) are \textit{sexist}, and the remaining $11,115$ ($68.6\%$) are \textit{clean}. The tweets have been authored by a total of $1,875$ unique users. Tweets in the \textit{racist} class come from $5$ of the users, while those in the \textit{sexist} class come from $527$ of them.

\section{Approach}
\subsection{Representing online communities}
We create two different graphs: the first one is identical to the community graph of Mishra et al. \shortcite{mishra} (referred to as the \textit{community} graph). It contains $1,875$ nodes representing each of the authors in the dataset. Two authors/nodes are connected by a single undirected edge if either one follows the other on Twitter. There are $453$ \textit{solitary authors} in the graph who are neither followed by nor follow any other author in the dataset. This graph is homogeneous, i.e., it has nodes (and hence edges) of a single type only.

Our second graph is an extended version of the first (referred to as the \textit{extended} graph) that additionally contains nodes representing the tweets of the authors. Specifically, in addition to the $1,875$ author nodes, the graph contains $16,202$ tweet nodes. Each tweet node is connected to a single author node, denoting that the tweet is elicited from that particular author. This graph is no longer homogeneous since it contains nodes and edges of two different types. 

\subsection{Generating author profiles}
\label{authprof}
We first describe the approach of Mishra et al. \shortcite{mishra} that learns author embeddings using \textit{node2vec} \cite{node2vec}; this serves as our baseline. We then move on to our semi-supervised approach based on graph convolutional networks \cite{gcn}.

\vspace{1 mm}
\noindent
\textbf{Node2vec.}
\textit{Node2vec} extends the \textit{word2vec} skip-gram model \cite{word2vec} to graphs in order to create low-dimensional embeddings for nodes based on their position and neighborhood. Specifically, for a given graph with nodes $V=\{v_1, v_2, \dots, v_n\}$, \textit{node2vec} aims to maximize the following log probability:
\begin{equation*}
\sum_{v \in V} \log \textrm{P}\,(N_s(v) \,|\, v)
\end{equation*}
where $N_s(v)$ denotes the neighbor set of node $v$ generated using neighbor sampling strategy $s$. The framework utilizes two different strategies for sampling neighbor sets of nodes: Depth-First Sampling (DFS) and Breadth-First Sampling (BFS). The former captures the structural role of nodes, while the latter captures the local neighborhood around them. Two hyper-parameters control the overall contribution of each of these strategies. Following Mishra et al. \shortcite{mishra}, we initialize these parameters to their default value of $1$ and set the embedding size and number of iterations to $200$ and $25$ respectively. Since \textit{node2vec} cannot produce embeddings for nodes without edges, we map the \textit{solitary authors} to a single zero embedding as done by Mishra et al.

\vspace{1 mm}
\noindent
\textbf{Graph convolutional networks.} We propose an approach for learning author profiles using \textsc{gcn}s applied to the \textit{extended} graph. In contrast to \textit{node2vec}, our method allows us to additionally propagate information with respect to whether tweets composed by authors and their communities are abusive or not. Specifically, as labels are available for a subset of nodes in our graph (i.e., the tweet nodes), we frame the task as a graph-based semi-supervised learning problem, allowing the model to distribute gradient information from the supervised loss on the labeled tweet nodes. This, in turn, allows us to create profiles for authors that not only capture the structural traits of their surrounding community but also their own linguistic behavior based on the types of tweets that they have composed.

We consider a graph $G = (V, E)$, where $V$ is the set of nodes ($|V| = n$) and $E$ is the set of edges. $A$ denotes the adjacency matrix of $G$. We assume that $A$ is symmetric ($A_{ij} = A_{ji}$), and that all nodes in $G$ have self loops ($A_{ii} = 1$). The significance of these assumptions is explained in Kipf and Welling \shortcite{gcn}. Let $D$ be the diagonal degree matrix defined as $D_{ii} = \sum_j A_{ij}$, and $F \in \mathbb{R}^{n \times m}$ be the input feature matrix that holds feature vectors of length $m$ for the nodes in $G$. We can now recursively define the computation that takes place at the $i^{th}$ \textit{convolutional layer} of a $k$-layer \textsc{gcn} as:

\begin{equation*}
O^{(i)} = \sigma\,(\,\widetilde{A}\,O^{(i-1)}\,W^{(i)}\,)
\end{equation*}

\noindent
with the computation at the first layer being:

\begin{equation*}
O^{(1)} = \sigma\,(\,\widetilde{A}\,F\,W^{(1)}\,)
\end{equation*}

Here, $\sigma$ denotes an activation function; $\widetilde{A} = D^{-\frac{1}{2}}A\,D^{-\frac{1}{2}}$ is the normalized adjacency matrix; $W^{(i)} \in \mathbb{R}^{d_{i-1}\times d_i}$ is the weight matrix of the $i^{th}$ convolutional layer; $O^{(i-1)} \in \mathbb{R}^{n\times d_{i-1}}$ represents the output from the preceding convolutional layer, where $d_i$ is the number of hidden units in the $i^{th}$ layer (note that $d_0 = m$, i.e., the length of the input feature vectors).

In our experiments, we apply a 2-layer \textsc{gcn} to the \textit{extended} graph.\footnote{Stacking more layers does not improve results on the validation set further.} Specifically, our \textsc{gcn} performs the following computation, yielding a softmax distribution over the $3$ classes in the dataset for each of the nodes:
\vspace{-5mm}

\begin{equation*}
O = softmax\,(\,\widetilde{A}\,\,ReLU\,(\,\widetilde{A}\,F\,W^{(1)}\,)\,W^{(2)})
\end{equation*}

\noindent
We set the input feature vectors in $F$ to be the binary bag-of-words representations of the nodes (following \citealt{gcn}); for author nodes, these representations are constructed over the entire set of their respective tweets. Note that $F$ is row-normalized prior to being fed to the \textsc{gcn}. We set the number of hidden units in the first convolutional layer to $200$ in order to extract $200$-dimensional embeddings for author nodes so that they are directly comparable with those from \textit{node2vec} . The number of hidden units in the second convolutional layer is set to $3$ for the output $O \in \mathbb{R}^{n\times 3}$ of the \textsc{gcn} to be a softmax distribution over the $3$ classes in the data.

The \textsc{gcn} is trained by minimizing the cross-entropy loss with respect to the labeled nodes of the graph. Once the model is trained, we extract $200$-dimensional embeddings $E = \widetilde{A}\,F\,W^{(1)}$ from the first layer (i.e., the layer's output without activation). This contains embeddings for author nodes as well as tweet nodes. For our experiments on author profiles, we make use of the former.

\subsection{Classification methods}
We experiment with five different supervised classification methods for tweets in the dataset. The first three (\textsc{lr}, \textsc{lr+auth}, \textsc{lr+extd}) serve as our baselines,\footnote{The implementations of the baselines are taken from \url{https://github.com/pushkarmishra/AuthorProfilingAbuseDetection}.} and the last two with \textsc{gcn}s\footnote{The code we use for our \textsc{gcn} models can be found at \url{https://github.com/tkipf/gcn}.} are the methods we propose.

\vspace{1 mm}
\noindent
\textbf{\textsc{lr}.} This method is adopted from Waseem and Hovy \shortcite{waseem_hovy} wherein they train a logistic regression classifier on character $n$-grams (up to $4$-grams) of the tweets. Character n-grams have been shown to be highly effective for abuse detection due to their robustness to spelling variations.

\vspace{1 mm}
\noindent
\textbf{\textsc{lr + auth}.} This is the state of the art method \cite{mishra} for the dataset we are using. For each tweet, the profile of its author (generated by \textit{node2vec} from the \textit{community} graph) is appended onto the tweet's character n-gram representation for training the \textsc{lr} classifier as above.

\vspace{1 mm}
\noindent
\textbf{\textsc{lr + extd}.} This method is identical to \textsc{lr + auth}, except that we now run \textit{node2vec} on the \textit{extended} graph to generate author profiles. Intuitively, since \textit{node2vec} treats both author and tweet nodes as the same and does not take into account the labels of tweets, the author profiles generated should exhibit the same properties as those generated from the \textit{community} graph.

\vspace{1 mm}
\noindent
\textbf{\textsc{gcn}.} Here, we simply assign a label to each tweet based on the highest score from the softmax distribution provided by our \textsc{gcn} model for the (tweet) nodes of the \textit{extended} graph.

\vspace{1 mm}
\noindent
\textbf{\textsc{lr + gcn}.} Identical to \textsc{lr + extd}, except that we replace the author profiles from \textit{node2vec} with those extracted by our \textsc{gcn} approach.

\begin{table*}[t]
\centering
\scalebox{1.0}{
\small
{\begin{tabular}{| c | c | c | c | c | c | c | c | c | c |}
\hline
\textbf{Method} &  \multicolumn{3}{c|}{\textbf{Racism}} & \multicolumn{3}{c|}{\textbf{Sexism}}  & \multicolumn{3}{c|}{\textbf{Overall}}\\ \hline
 & \textbf{\textsc{p}} & \textbf{\textsc{r}} & \textbf{\textsc{f}$_1$} & \textbf{\textsc{p}} & \textbf{\textsc{r}} & \textbf{\textsc{f}$_1$} & \textbf{\textsc{p}} & \textbf{\textsc{r}} & \textbf{\textsc{f}$_1$}\\ \hline
\textsc{lr} & \textbf{80.59} & 70.62 & 75.28 & 83.12 & 62.54 & 71.38 & 83.18 & 75.62 & 78.75\\
\textsc{lr + auth} & 77.95 & 78.35 & 78.15 & 87.28 & 78.41 & 82.61 &  85.26 & 83.28 & 84.18\\
\textsc{lr + extd} & 77.95 & 78.35 & 78.15 & 87.02 & 78.73 & 82.67 & 85.17 & 83.33 & 84.17\\ \hline
\textsc{gcn}$^\dagger$ & 74.12 & 64.95 & 69.23 & 82.48 & \textbf{82.22} & 82.35 & 81.90 & 79.42 & 80.56\\
\textsc{lr + gcn}$^\dagger$ & 79.08 & \textbf{79.90} & \textbf{79.49} & \textbf{88.24} & 80.95 & \textbf{84.44} & \textbf{86.23} & \textbf{84.73} & \textbf{85.42}\\ \hline
\end{tabular}}}
\caption{The baselines (\textsc{lr}, \textsc{lr + auth/extd}) vs. our \textsc{gcn} approaches ($^\dagger$) on the racism and sexism classes. \textit{Overall} shows the macro-averaged metrics computed over the $3$ classes: \textit{sexism}, \textit{racism}, and \textit{clean}.}
\label{res_abusive}
\end{table*}

\section{Experiments and results}
\subsection{Experimental setup}
We run every method $10$ times with random initializations and stratified train--test splits. Specifically, in each run, the dataset is split into a randomly-sampled train set ($90\%$) and test set ($10\%$) with identical distributions of the $3$ classes in each. In methods involving our \textsc{gcn}, a small part of the train set is held out as validation data to prevent over-fitting using \textit{early-stopping} regularization. When training the \textsc{gcn}, we only have labeled tweet nodes for those tweets in the \textit{extended} graph that are part of the train set. Our \textsc{gcn} is trained using the parameters from the original paper \cite{gcn}: \textit{Glorot} initialization \cite{glorot}, \textsc{adam} optimizer \cite{adam} with a learning rate of $0.01$, \textit{dropout} regularization \cite{dropout} rate of $0.5$, $200$ training epochs with an early-stopping patience of $10$ epochs.

\subsection{Results and analysis}
In Table \ref{res_abusive}, we report the mean precision, recall, and F$_1$ on the \textit{racism} and \textit{sexism} classes over the $10$ runs. We further report the mean macro-averaged precision, recall, and F$_1$ for each method (`Overall') to investigate their overall performance on the data. \textsc{lr + gcn} significantly ($p < 0.05$ on paired t-test) outperforms all other methods. The author profiles from \textit{node2vec} only capture the structural and community information of the authors; however, those from the \textsc{gcn} also take into account the (abusive) nature of the tweets composed by the authors. As a result, tweets like ``\textit{\#MKR \#mkr2015 Who is gonna win the peoples choice?}'' that are misclassified as sexist by \textsc{lr + auth} (because their author is surrounded by others producing sexist tweets) are correctly classified as clean by \textsc{lr + gcn}.

\textsc{gcn} on its own achieves a high performance, particularly on the \textit{sexism} class where its performance is typical of a community-based profiling approach, i.e., high recall at the expense of precision. However, on the \textit{racism} class, its recall is hindered by the same factor that Mishra et al. \shortcite{mishra} highlighted for their \textit{node2vec}-only method, i.e., that racist tweets come from 5 unique authors only who have also contributed sexist or clean tweets. The racist activity of these authors is therefore eclipsed, leading to misclassifications of their tweets. \textsc{lr + gcn} alleviates this problem by incorporating character n-gram representations of the tweets, hence not relying solely on the linguistic behavior of their authors.

\begin{figure}[ht!]
\centering
\subfloat[Author profiles from the \textit{community} graph]{
\hspace*{26pt}
\includegraphics[width=5cm]{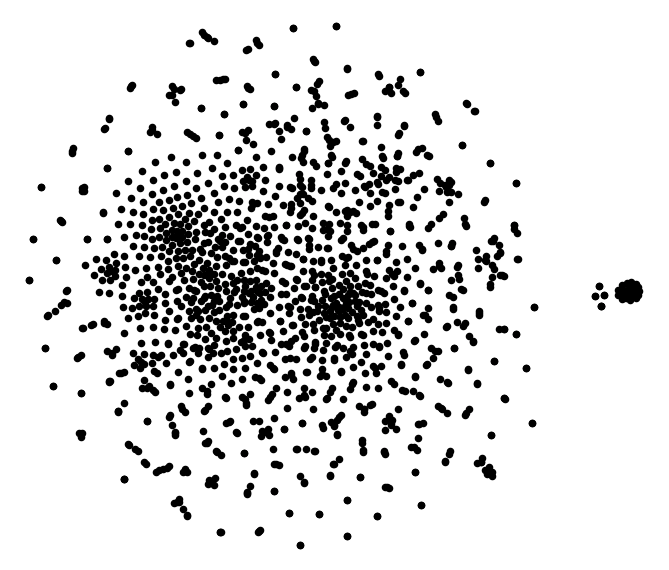}
\hspace*{26pt}
}
\qquad
\subfloat[Author profiles from the \textit{extended} graph]{
\hspace*{26pt}
\includegraphics[width=5cm]{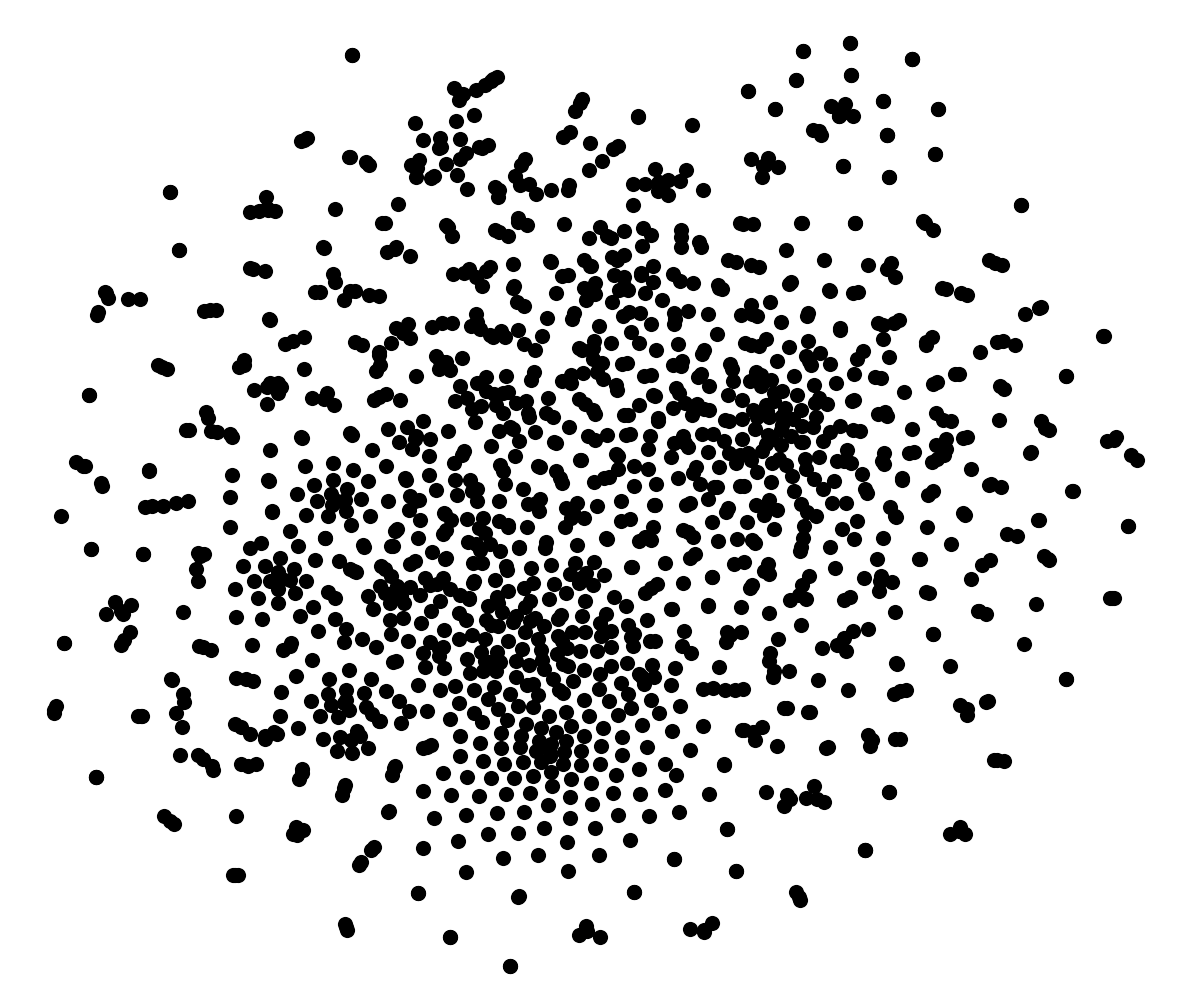}
\hspace*{26pt}
}
\caption{Visualizations of the \textit{node2vec} author profiles from the \textit{community} and \textit{extended} graphs.}
\label{tsne_node2vec}
\end{figure}

Figure \ref{tsne_node2vec} shows the t-\textsc{sne} \cite{tsne} visualizations of \textit{node2vec} author profiles from the \textit{community} and \textit{extended} graphs. Both visualizations show that some authors belong to densely-connected communities while others are part of more sparse ones. The results from \textsc{lr + auth} and \textsc{lr + extd} have insignificant differences, further confirming that their author profiles have similar properties. In essence, \textit{node2vec} is unable to gain anything more from the \textit{extended} graph than what it does from the \textit{community} graph.

\begin{figure}[ht!]
\centering
\includegraphics[width=5cm]{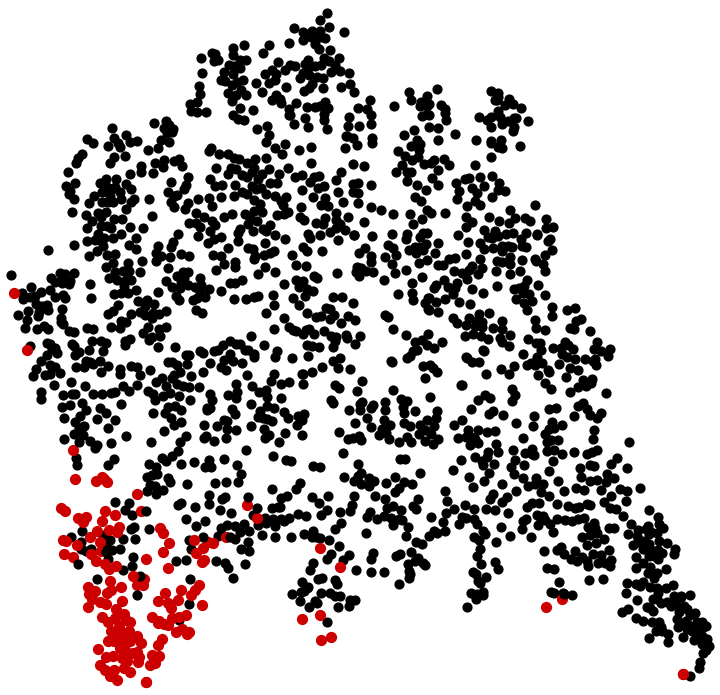}
\caption{Visualization of the author profiles extracted from our \textsc{gcn}. Red dots represent the authors who are deemed abusive (racist or sexist) by the \textsc{gcn}.}
\label{gcn_embeds}
\end{figure}

Figure \ref{gcn_embeds} shows a t-\textsc{sne} visualization of the author profiles generated using our \textsc{gcn} approach. Red dots denote the authors who are abusive (sexist or racist) according to our model (i.e., as per the softmax outputs for the author nodes).\footnote{Note that there are no such gold labels for authors in the dataset itself.} The red dots are mostly clustered in a small portion of the visualization, which corroborates the notion of homophily amongst abusive authors.

Despite the addition of improved author profiles, several abusive tweets remain misclassified. As per our analysis, many of these tend to contain \textsc{url}s to abusive content but not the content itself, e.g., ``\textit{@MENTION: Logic in the world of Islam http://t.co/6nALv2HPc3}" and ``\textit{@MENTION Yes. http://t.co/ixbt0uc7HN}". Since Twitter shortens all \textsc{url}s into a standard format, there is no indication of what they refer to. One possible way to address this limitation could be to append the content of the \textsc{url} to the tweet; however this can lead to misclassifications in cases where the tweet is disagreeing with the \textsc{url}.
Another factor in misclassifications is the deliberate obfuscation of words and phrases by authors in order to evade detection, e.g., ``\textit{Kat, a massive c*nt. The biggest ever on \#mkr \#cuntandandre}". Mishra et al. \shortcite{mishra2} demonstrate in their work that character-based word composition models can be useful in dealing with this aspect.

\section{Conclusions}
In this paper, we built on the work of Mishra et al. \shortcite{mishra} that introduces community-based profiling of authors for abusive language detection. We proposed an approach based on graph convolutional networks to show that author profiles that directly capture the linguistic behavior of authors along with the structural traits of their community significantly advance the current state of the art.

\section*{Acknowledgments}

We would like to thank the anonymous reviewers for their useful feedback. Helen Yannakoudakis was supported by Cambridge Assessment, University of Cambridge.

\bibliography{naaclhlt2019}
\bibliographystyle{acl_natbib}
\end{document}